# Deep Air Quality Forecasting Using Hybrid Deep Learning Framework

Shengdong Du, Tianrui Li, Senior Member, IEEE, Yan Yang, Member, IEEE, and Shi-Jinn Horng

**Abstract**—Air quality forecasting has been regarded as the key problem of air pollution early warning and control management. In this paper, we propose a novel deep learning model for air quality (mainly PM2.5) forecasting, which learns the spatial-temporal correlation features and interdependence of multivariate air quality related time series data by hybrid deep learning architecture. Due to the nonlinear and dynamic characteristics of multivariate air quality time series data, the base modules of our model include one-dimensional Convolutional Neural Networks (1D-CNNs) and Bi-directional Long Short-term Memory networks (Bi-LSTM). The former is to extract the local trend features and spatial correlation features, and the latter is to learn spatial-temporal dependencies. Then we design a jointly hybrid deep learning framework based on one-dimensional CNNs and Bi-LSTM for shared representation features learning of multivariate air quality related time series data. We conduct extensive experimental evaluations using two real-world datasets, and the results show that our model is capable of dealing with PM2.5 air pollution forecasting with satisfied accuracy.

✦ ────────── ─ ──────────

## 1 INTRODUCTION

WITH the acceleration of industrialization and the rapid development of urbanization, the problem of urban air pollution has become more and more serious, which has badly affected our living environment and physical health. Therefore, research on air quality forecasting is very important and has always been regarded as a key issue in environmental protection. It is also an important means to guide the scientific decision-making of severe air pollution warning and air pollution control. Many large cities have established air quality monitoring stations to detect the city's PM2.5 and other air pollutants in real time. Early diagnosis of air pollution occurrence and PM2.5 concentration value evolution is considered to be a key problem of air quality forecasting task.

In recent years, some researchers have made efforts on air pollution occurrence and air quality forecasting [1] [2]. However, most of these studies do rely on mathematical equations or simulation techniques to describe the evolution of air pollution [3]. These traditional methods are represented by classic shallow machine learning algorithms. For example, Dong et al. presented a novel approach which is based on hidden semi-Markov models (HSMMs) for PM2.5 concentration value prediction [4]. Donnelly et al. proposed a model for producing real-time air quality forecasts with both high accuracy and high computational efficiency based on Integrated Parametric and Nonparametric Regression method [5]. Because air pollution is usually affected by weather, traffic and other factors, it is difficult to accurately represent and predict by the statistical methods and shallow machine learning models.

In the big data era, with the rapid development and application of the Internet of Things and sensor technology, air quality forecasting is increasingly dependent on a variety of sensors and related data acquisition equipment to collect the urban air big data, e.g. PM2.5, NO2, PM10, weather condition data and traffic data, etc. Since traditional shallow learning models still have bottlenecks in handling big data, new air quality forecasting methods need data-driven model support [7][8]. Deep learning is currently the most popular data-driven method [9], which can extract and learn the inherent features of various air quality data automatically. Since 2012, deep learning has made great progress in research and applications of image processing, audio processing, and natural language understanding [10][11][12]. Although air quality forecasting task usually adopts the traditional shallow machine learning methods, the deep learning method for time series analysis and air quality prediction is getting more and more attention from researchers [13][14][34][36]. In the issue of air quality forecasting, which is a typical multivariate time series analysis problem [35], it's a useful exploration of learning various implicit features and long temporal dependencies of multivariate air quality time series data based on the hybrid deep learning model.

In this paper, we propose an end-to-end model for the air quality forecasting problem in one framework called the Deep Air Quality Forecasting Framework (DAQFF), which addresses the dynamic, spatial-temporal and nonlinear characteristics of multivariate air quality time series

_________

- *Shengdong Du, Tianrui Li, Yan Yang are with the School of Information Science and Technology, Southwest Jiaotong University. Emails: {sddu, trli,yyang}@swjtu.edu.cn.*
- *Shi-Jinn Horng is with the Department of Computer Science and Information Engineering, National Taiwan University of Science and Technology, is also with School of Computer Science, University of Technology Sydney. Email: horngsj@yahoo.com.tw.*
- *Corresponding authors: Tianrui Li and Shi-Jinn Horng*

data by a hybrid deep learning model. The proposed model can learn the local trend pattern and long spatilatemporal dependencies of multivariate air quality related time series data, e.g. PM2.5, wind speed, temperature, etc. It is also shown that the proposed model DAQFF has good forecasting performance and generalization ability. Experiments indicate that our proposed method is effective in air quality prediction tasks.

The rest of the paper is organized as follows: Section II presents the related works. Section III shows an overview of the deep air quality forecasting framework, including the overall design of our model, e.g. how to expand and assemble basic deep neural network modules into our model. Section IV describes the comparative experiments, and the effectiveness of the proposed framework is analyzed and evaluated. We draw conclusions and directions for future research in the last section.

## 2 RELATED WORKS

Air quality forecasting has a good study history in the literature, most of the existing works solve the problems of air quality forecasting using statistical methods and shallow machine learning models [3] including Regression [5], ARIMA [17], HMM [4], and Artificial Neural Network [16]. Zhang et al. presented a comprehensive assessment of the history, current status, major research and future directions of real-time air quality forecasting problems [1] [2]. Zhou et al. proposed a probabilistic dynamic causal (PDC) model based on Lasso-Granger to uncover the dynamic temporal dependencies of PM2.5 [6]. Zhou et al. developed a hybrid model for one-day-ahead PM2.5 forecasting based on ensemble empirical mode decomposition and a general regression neural network method [16]. Deleawe et al. investigated the use of machine learning technologies to predict the CO2 level, which is an indicator of air quality in urban air environments [15].

In recent years, air quality forecasting based on big data analysis has become a research hotspot. Because air quality related time series data have dynamic and nonlinear characteristics, more and more researchers are trying to use data-driven models, especially in the field of urban computing [18]. A large number of air quality forecasting methods based on the big data have been proposed to help air pollution warning and control [37]. Zheng et al. developed a semi-supervised learning approach for air quality forecasting which is based on a co-training framework consisting of two separated classifiers (ANN and CRF) [7]. Hsieh et al. presented a novel method which can infer the real-time and fine-grained air quality throughout a city by a semi-supervised inference model [8]. Zheng et al. also proposed a real-time air quality forecasting framework which uses data-driven models to predict fine-grained air quality [19].

More recently, deep learning has been widely applied to sequence data processing and time series problems [20][21][24]. Air quality is typical time series data. Li et al. presented a novel spatial-temporal deep learning (STDL)-based air quality prediction method which inherently considers spatial and temporal correlations [22]. Ong et al. proposed a deep recurrent neural network (DRNN) for air pollution prediction which is improved by using the autoencoder model as a novel pre-training method [23]. Qi et al. developed a general and effective approach to solve interpolation, prediction and feature analysis in one model which is called Deep Air Learning (DAL) [14]. Moreover, the deep convolution network could process time series features of citywide crowd big data and Zhang et al. proposed a novel deep residuals network to analyze how the congestions are evolving [25]. The hybrid deep learning method is based on the idea of a combination of various deep neural network structures and has achieved good application effects in many fields, e.g. face detection and video classification [26][27], but it has not yet been well applied for air quality forecasting problems.

In this paper, by a comparison of traditional shallow machine learning models and classic deep learning models, we propose a new end-to-end air quality forecasting framework, DAQFF, based on the hybrid deep learning method, which is motivated to address local trend features and long temporal dependency problems by utilizing the multivariate time series data and performing feature selection automatically. The proposed DAQFF can extract and learn the nonlinear spatial-temporal features of air quality related time series data under different conditions such as different weather conditions and different traffic states.

## 3 METHODOLOGY

### 3.1 Problems and Motivations

Air quality forecasting has been a key issue in early warning and control of urban air pollution. Its goal is to anticipate changes in the PM2.5 value of air pollution at observation points over time. The observation time period is usually set for one hour, which is decided by the ground-based air-quality monitoring station. Typical air pollution data, e.g. PM2.5, is shown in Fig. 1.

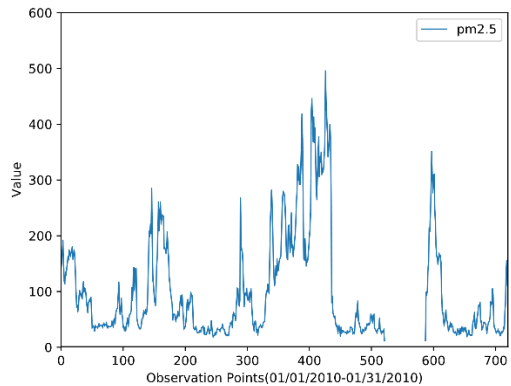

Fig. 1. PM2.5 values in one month (01/01/2010-01/31/2010) of Beijing air pollution data set from UCI [31].

PM2.5 prediction problem is illustrated as follows. Given time $T$, the prediction task is to anticipate the PM2.5 concentration value $P_{i,T+1}$ at time $T+1$ or $P_{i,T+n}$ at time

$T + n$ which models the history air quality related time series dataset $AQD = \{AQD_{i,t} | i \in O, t = 1,2, ... , T$ in the past$\}$, where $AQD$ represents the history air quality related data, $O$ means the overall observation points, and $AQD$ not only includes PM2.5 itself but also includes other air quality related time series data such as press, temperature, wind speed, etc. Fig. 2 shows an example.

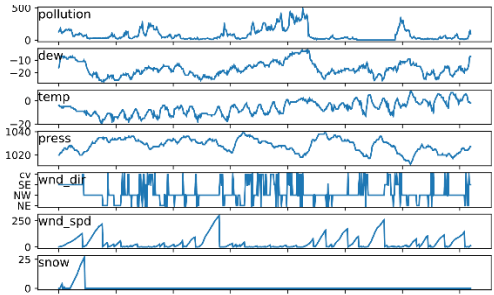

Fig. 2. Air quality related time series data in one month (01/01/2010-01/31/2010) (include PM2.5 pollution concentration, temperature, pressure, wind speed, wind direction, snow, rain, etc.) of Beijing PM2.5 data set from UCI [31].

As shown in Fig. 2, air quality data usually contains the real-valued PM2.5 pollutant, and some other datasets also have CO2 and PM10, etc. In addition to pollutant data, air quality is highly related by meteorological observation data. For example, high wind speed will reduce the concentration of PM2.5, high humidity usually aggravates air pollution, and high atmospheric pressure usually results in good air quality [7][19]. Therefore, the above data characteristics are very important for air quality forecasting task.

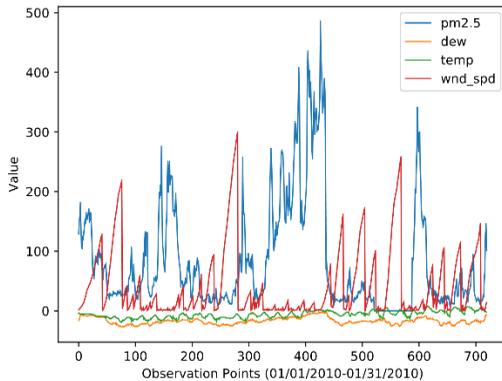

Fig. 3. The interdependences and correlations of multivariate air quality time series data (such as PM2.5, dew point, temperature, wind speed etc.).

How to process and capture the spatial-temporal features of above air quality data items is the key point for air quality forecasting. Taking the PM2.5 data itself as an example (See Fig. 3, for a month observation data points during 01/01/2010-01/31/2010), there is contextual information among the observation points in the PM2.5 and wind speed time series, and the historical state has some influence on the evolution of future trends. That is to say, the adjacent data points and the periodic interval of the air quality time series data usually have a strong correlation with each other.

In addition, air quality data have sharp nonlinearities resulting from transitions from bad air pollution to good air quality and vice versa. Air quality forecasting task is challenging due to rapidly changing weather and pollutant emission conditions and it is influenced by a lot of factors. Moreover, these factors are nonlinear and dynamic (See Fig. 3), such as wind speed, temperature, humidity, and pollutants itself. Those influences are complex and highly non-linear and it is hard to precise forecasting air quality for a specific time and place. Because these factors are inherently interdependent, how to deal with the interdependence and exploit it from the multivariate air quality related time series data is another key problem for air quality forecasting.

Regarding the issues above, an air quality forecasting (mainly predicting PM2.5) method based on a hybrid deep learning architecture is proposed in this paper. In general, because the statistical characteristics of air quality related time series data are different (different time series always have different representations and related structures), it is difficult to use shallow machine learning models for fusion modeling. Many researchers have studied the hybrid deep learning model, which is usually effective for improving the performance of classic deep learning models [26].

CNN is very popular for image processing and target recognition [10], and it is also successfully applied to time series forecasting tasks [24], due to the one-dimensional structure of single time series and two-dimensional structure of multivariate time series. For the above reasons, researches on target recognition in images also can be applied to time series modeling as well. Meanwhile, recurrent neural networks (RNN) model can be used for temporal representation learning of the long dependency features. Because a feedback loop is created in the internal state of the RNN network [12], this is why RNN performs better at predicting time series. LSTM is a special kind of RNN, capable of learning long-term dependencies. We use a bi-directional LSTM to process time-series information in two directions with two separate hidden layers and then feed this information to the same output layer [29][30] so that it can access both past and future contexts for each point in the time series.

### 3.2 Overview of the Deep Air Quality Forecasting Framework

In the following, we describe the air quality forecasting framework, DAQFF, based on the hybrid deep learning architecture. It is a combination of multiple one-dimensional CNNs and Bi-directional LSTM that take into account the spatial-temporal dependence of air quality-related time series data. Because there have correlations between local trend features and long dependencies of air quality multivariate time series data, PM2.5 time series is also related to other air quality time series data. And these factors are inherently interdependent. Fig. 4 shows the graphical illustration of the deep air quality forecasting framework. From Fig. 4, the overall model consists of two main components: one is the multiple convolution layers (one-dimensional CNNs) for local trend and spatial correlation features learning of time series data and the other is the bi-directional LSTM for getting the long dependency temporal features from the corresponding time series data.

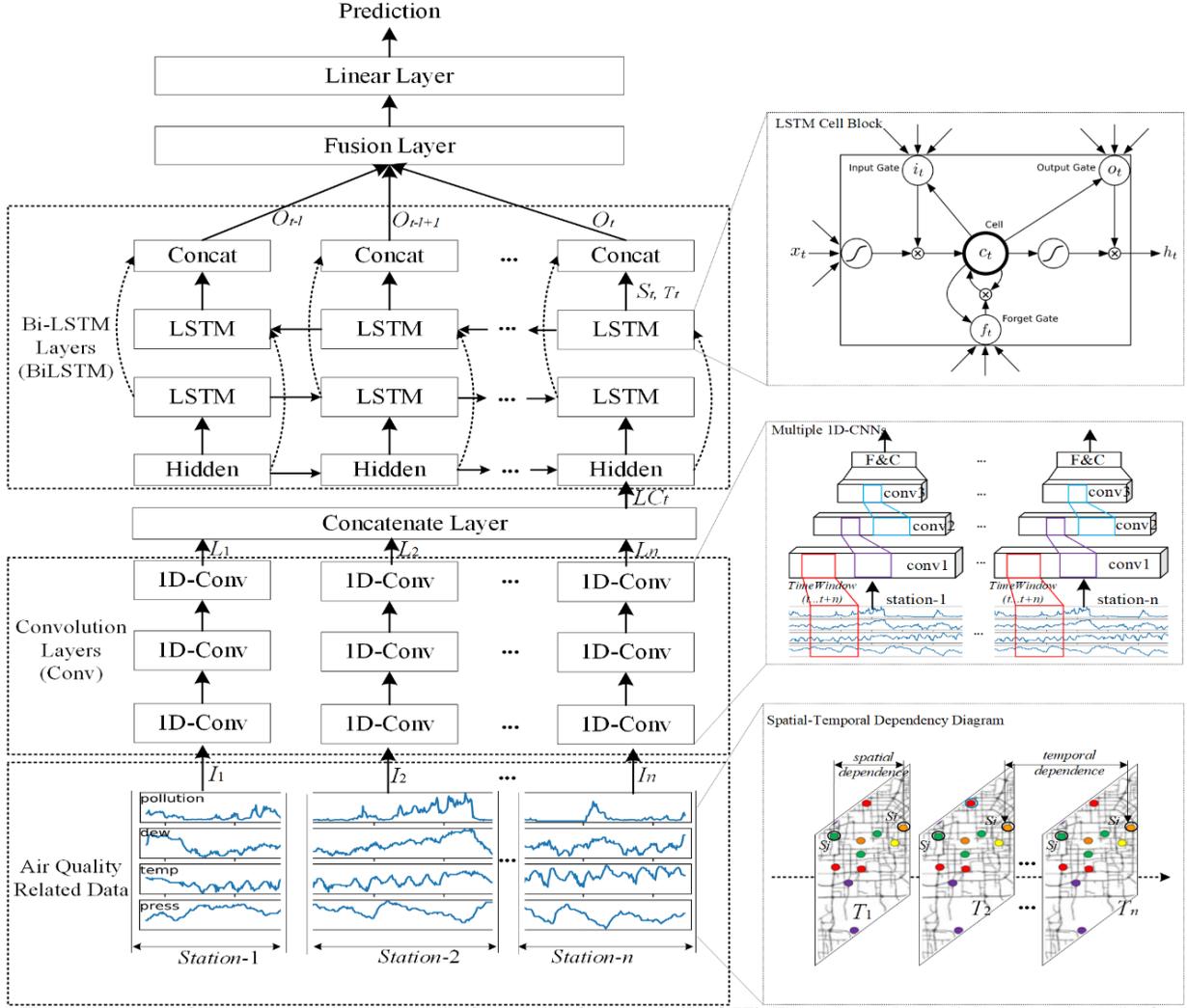

Fig. 4. The architecture of the proposed deep air quality forecasting framework (DAQFF). The proposed model works on air quality forecasting using hierarchical feature representation learning and multi-scale spatial-temporal dependency feature fusion learning.

To exploit spatial-temporal dependency features of different air quality related time series data (see the lower right corner of Fig. 4), the first step is to train multiple one-dimensional CNNs to extract local trend features and possible spatial correlation features of multiple stations time series data. Unlike traditional image processing methods (which are fed with two-dimensional image pixels), the inputs to our DAQFF model are multiple one-dimensional time series. We employ an improved CNN model, which can compress the length of air quality time series. Rather than learning the features of each single time series separately, we learn all the time series data of each observation point of multiple stations.

Then, the extracted features (including the local trend features of each station data and the possible spatial correlation features of multiple stations data) of many one-dimensional CNNs are concatenated and fed into certain bidirectional LSTMs. These LSTMs learns spatial-temporal dependency features from both past and future contexts utilizing time series in forward and backward directions simultaneously.

Given an air quality time series dataset $I_i$ of a station ($i$ denotes the station number), the process to learn the spatial-temporal dependency features of multiple stations data can be represented as follows:

$$Convs(I_i) \rightarrow L_i \quad (1)$$
$$Concatenate(L_1 \dots L_i \dots L_n) \rightarrow LC_t \quad (2)$$
$$BiLSTM(LC_t) \rightarrow S_t, T_t \quad (3)$$
$$Concat(S_t, T_t) \rightarrow O_t \quad (4)$$

where $L_i$ denotes the local trend features of single station time series data $I_i$, and $LC_t$ denotes the concatenated local trend features of all stations and the hidden spatial correlation features between all stations. These spatial correlation features with local trend features are concatenated and learned by the Bi-LSTM model automatically. Note that $S_t$ and $T_t$ denote the spatial and temporal dependency features, respectively, which are extracted from multiple stations data, and $Concat(S_t, T_t)$ represents the feature level fusion result. $O_t$ is the shared representation between $S_t$ and $T_t$. Next, we use a fusion layer to concatenate all the spatial-temporal shared features among different time series data together. The model is formulated as follows:

$$F((O_{t-l}, O_{t-l+1}, ..., O_t), W^i, b^i) \rightarrow M_\pi, \quad i=1, 2, ..., n \quad (5)$$

where $M_\pi$ denotes the joint fusion representation for different learned spatial-temporal dependency features which are extracted from multiple air quality time series data. $W^i$ and $b^i$ are weights and biases, respectively, which are learned by the fusion model with all training datasets, where $i$ indicates the $i^{th}$ time window of input time series data, $l$ indicates the time window size (also called lookup size). The training objective function of DAQFF model is as follows:

$$\operatorname*{argmin}_{\theta} C_i = \frac{1}{n} \sum_{i=1}^{n} \sum_{j=1}^{m} ||\hat{y}_i^j - y_i^j||^2 \quad (6)$$

The final model training problem is to minimize the overall error $C_i$ of training samples for each time window time series, where $i$ indicates each time window time series input ($i=1, 2, ..., n$), $j$ indicates the input samples number of a time window input data ($j=1, 2, ..., m$) and $\theta$ is the parameter space including $W_l^i$ and $b_l^i$ of each layer.

Based on the above process, one-dimensional CNNs are used to extract the local trend and spatial correlation features. Bi-LSTM is used to capture and learn the spatial-temporal dependency features of the sequence and obtains the correlation pattern of the time series context. Then we fuse these learned spatial-temporal dependency features by concatenating layers. Finally, we input these joint fusion features into the linear regression layer for final prediction. In this way, DAQFF combines multiple one-dimensional CNNs and bi-directional LSTM in one end-to-end deep learning architecture, which can simultaneously extract the local trend features and the spatial-temporal dependency features of air quality related multivariate time series data.

### 3.3 Multiple 1D-CNNs for local trend and spatial features learning

CNN not only has excellent performance in image processing [10], but also can be effectively applied on time series data mining. A typical CNN has three layers: convolutional layer, activation layer, and pooling layer. Unlike the classical CNN model (also traditional two-dimensional CNN used for images), we propose to use multiple one-dimensional filters convolved (1D-CNNs) over all time steps of air quality time series data. The computing processes of 1D-CNN layers are formulated as below:

$$c_j^l = \sum_i x_i^{l-1} * w_{ij}^l + b_j^l \quad (7)$$
$$x_j^l = ReLU(c_j^l) \quad (8)$$
$$x_j^l = Flatten(x_j^l) \quad (9)$$
$$x_k^{l+1} = FC(w_{kj}^{l+1} x_j^l + b_k^{l+1}) \quad (10)$$

Note that Eq. (7) and Eq. (8) model the convolutional layer learning process, where * denotes a convolution operator, $w_{ij}^l$ and $b_j^l$ are the filters and biases, respectively.

We use ReLU as the activation function. $x_i^{l-1}$ and $c_j^l$ represent the input and output vectors to a convolution layer, respectively. Here $l$ represents the involved layer. We use three convolution layers for local trend feature learning. Each layer learns a non-linear representation from the previous layer, and the learned representation is then fed into the next layer to form hierarchical feature representations. After processing three convolution layer, we use a flatten layer to transform the high-level representation to a feature vector and use a fully connected layer to reduce the dimension of the final output vector.

As introduced above, the multi-station input air quality time series data are processed using multiple 1D-CNNs, and are flattened into the fully connected layer. Then the final output is given by a concatenated layer, which not only captures the local trend features of single station time series data (as one dimensional filter is used in each convolutional layer, the local trend change features of the time series over time can be captured.), but also integrates the possible spatial correlation features of multiple stations.

Moreover, one-dimensional CNN's local perception and weighted sharing features can reduce the number of parameters for processing multivariate time series data, thereby improving learning efficiency. Thus, our method can learn more deep representation features of air quality related data.

### 3.4 Bi-LSTM for long temporal dependencies learning

Although traditional statistical methods like ARIMA and shallow learning models similar to deep neural networks can process time series, the efficiency is not so good, because it does not take into account the long-term temporal dependence of time series data. In order to overcome this shortcoming, Long Short-term Memory network (LSTM) is a good option [32], which is a popular dynamic model for handling sequence tasks.

As shown in the upper right corner of Fig. 4, the LSTM Cell Block represents a typical LSTM diagram [33]. The memory cell of each LSTM block contains four main components. The collaboration of these components enables cells to learn and memory long dependency features. The typical LSTM block computing process is as follows:

$$i_t = \sigma(U^{(i)} x_t + W^{(i)} h_{t-1} + b_i) \quad (11)$$
$$f_t = \sigma(U^{(f)} x_t + W^{(f)} h_{t-1} + b_f) \quad (12)$$
$$o_t = \sigma(U^{(o)} x_t + W^{(o)} h_{t-1} + b_o) \quad (13)$$
$$\tilde{s}_t = tanh(U^{(c)} x_t + W^{(c)} h_{t-1} + b_c) \quad (14)$$
$$s_t = f_t \circ s_{t-1} + i_t \circ \tilde{s}_t \quad (15)$$
$$h_t = o_t \circ tanh(s_t) \quad (16)$$

As shown in the above formulas, $i_t$ represents the input gate and it decides the new information input the memory cell. $f_t$ represents the forget gate which decides how much information should be discarded. $o_t$ indicates the output gate which decides the amount of information should transfer to the next time step or to the output. $\tilde{s}_t$ is a neuron with a self-recurrent cell like RNN. $s_t$ is the internal memory cell of LSTM block which is summed by two parts. The first part is calculated by the previous internal memory state $s_{t-1}$ and forget gate $f_t$. The second part is calculated by element wise multiplication of self-recurrent state $\tilde{s}_t$ and input gate $i_t$. $h_t$ is hidden state of LSTM block.

One disadvantage of traditional LSTMs is that they can only utilize the previous context of sequence data, and Bi-directional LSTM can process the time series data in two directions simultaneously through two independent hidden layers [30], and these data are concatenated and fed forward to the output layer. In other words, Bi-directional LSTM processes the time series data in two directions iteratively (forward layer from $t = 1$ to $T$, backward layer from $t = T$ to 1).

$$\vec{i}_t = \sigma(\vec{U}^{(i)}\vec{x}_t + \vec{W}^{(i)}\vec{h}_{t-1} + \vec{b}_i) \quad (17)$$
$$\vec{f}_t = \sigma(\vec{U}^{(f)}\vec{x}_t + \vec{W}^{(f)}\vec{h}_{t-1} + \vec{b}_f) \quad (18)$$
$$\vec{o}_t = \sigma(\vec{U}^{(o)}\vec{x}_t + \vec{W}^{(o)}\vec{h}_{t-1} + \vec{b}_o) \quad (19)$$
$$\vec{\tilde{s}}_t = tanh(\vec{U}^{(c)}\vec{x}_t + \vec{W}^{(c)}\vec{h}_{t-1} + \vec{b}_c) \quad (20)$$
$$\vec{s}_t = \vec{f}_t \circ \vec{s}_{t-1} + \vec{i}_t \circ \vec{\tilde{s}}_t \quad (21)$$
$$\vec{h}_t = \vec{o}_t \circ tanh(\vec{s}_t) \quad (22)$$

$$\overleftarrow{i}_t = \sigma(\overleftarrow{U}^{(i)}\overleftarrow{x}_t + \overleftarrow{W}^{(i)}\overleftarrow{h}_{t-1} + \overleftarrow{b}_i) \quad (23)$$
$$\overleftarrow{f}_t = \sigma(\overleftarrow{U}^{(f)}\overleftarrow{x}_t + \overleftarrow{W}^{(f)}\overleftarrow{h}_{t-1} + \overleftarrow{b}_f) \quad (24)$$
$$\overleftarrow{o}_t = \sigma(\overleftarrow{U}^{(o)}\overleftarrow{x}_t + \overleftarrow{W}^{(o)}\overleftarrow{h}_{t-1} + \overleftarrow{b}_o) \quad (25)$$
$$\overleftarrow{\tilde{s}}_t = tanh(\overleftarrow{U}^{(c)}\overleftarrow{x}_t + \overleftarrow{W}^{(c)}\overleftarrow{h}_{t-1} + \overleftarrow{b}_c) \quad (26)$$
$$\overleftarrow{s}_t = \overleftarrow{f}_t \circ \overleftarrow{s}_{t-1} + \overleftarrow{i}_t \circ \overleftarrow{\tilde{s}}_t \quad (27)$$
$$\overleftarrow{h}_t = \overleftarrow{o}_t \circ tanh(\overleftarrow{s}_t) \quad (28)$$

$$h_t = \vec{h}_t \circ \overleftarrow{h}_t \quad (29)$$

The above equations show the layer functions of Bi-LSTM, and the two direction arrows denote the forward and backward process, respectively. $h_t$ represents the final hidden element of Bi-LSTM, which is the concatenated vector of the forward output $\vec{h}_t$ and the backward output $\overleftarrow{h}_t$. Through the above process, Bi-LSTM can learn both past and future features of time series data and the predictive output is generated from past and future contexts.

## 4 EXPERIMENTS

In this section, we use two real air quality data sets to conduct experiments to analyze and evaluate the performance of the proposed model. Through the comparison of classical shallow learning models, baseline deep learning models and our model DAQFF, the forecasting performance and effectiveness of the proposed model are validated.

### 4.1 Datasets

Our experiment uses two real-world air quality datasets: The first one is the Beijing air quality dataset from UCI [31], which includes meteorological data and PM2.5 pollution data. The dataset is collected every hour and is sourced from the data interface released by the US Embassy in Beijing [28]. As an experimental air quality UCI data set, it contains different attributes such as date, time, temperature, humidity, wind speed, wind direction, and PM2.5 values. And the second dataset is the Urban Air Quality Dataset collected in the Urban Air project of Microsoft Research [19]. The details of the two experimental data set are listed as follows (as shown in Table 1):

TABLE 1
EXPERIMENTS DATASETS DESCRIPTION

| Dataset | Beijing PM2.5 Dataset[31] | Urban Air Quality Dataset [19] |
|---|---|---|
| Data type | multivariable time series | multivariable time series |
| Intervals | 60-minutes | 60-minutes |
| Location | Beijing | Beijing |
| Time Span | 01/01/2010-12/31/2014 | 05/01/2014-04/30/2015 |
| Variable number | 8 | 14 |
| Used records | 43,824 | 278,023 |
| Station number | 1 station | 36 stations |

**Beijing PM2.5 Dataset**: This hourly dataset contains the PM2.5 data of the US Embassy in Beijing. Meanwhile, meteorological data are also included. Data items include PM2.5 concentration, Dew Point, Temperature, Pressure, Combined wind direction, Cumulated wind speed (m/s), Cumulated hours of snow, Cumulated hours of rain. The dataset used for experiments is ranged from 01/01/2010 to 12/31/2014, which has 43824 records.

**Urban Air Quality Dataset**: This hourly dataset is comprised of six parts of data over a period of one year (from 05/01/2014 to 04/30/2015), which has been used in [7] [19] to infer the fine-grained air quality of current and future times. We select the data from Beijing as the experimental dataset, which contains a total of 278,023 records from 36 monitoring stations, where the data items include PM2.5, PM10, NO2, CO, O3, SO2, weather, temperature, humidity, pressure, wind speed and wind direction, etc.

### 4.2 Experimental Setup

This section describes the hardware and software environment of the experiment and the configuration of relevant parameters. The open source deep learning library Keras which based on Tensorflow is used to build baseline deep learning models and DAQFF model, and Scikit-learn is used to build shallow learning models. All experiments are conducted on a PC Server, and the server configuration is Intel(R) Xeon(R) CPU E5-2623 3.00GHz, 4 GPUs each is 12G NVIDIA Tesla K80C, and memory is 128GB.

Our framework is compared with two classic shallow learning models and five baseline deep learning models. They are summarized as follows.

ARIMA is one of the most common traditional statistical methods in time series prediction.

SVR (Support Vector Regression) is a kernel method of machine learning which also can be used for time series forecasting. And the kernel-based SVR can make it possible to learn nonlinear trend of the training dataset. There are three SVR models with different kernels (RBF, poly and linear).

RNN (Recurrent Neural Network) is a popular deep learning method for handling sequence tasks. GRU (Gated Recurrent Units) and LSTM (Long Short-term Memory) are the most popular variants of RNN. CNN (Convolutional Neural Network) is widely used in image processing, but one-dimensional CNN can also be used for time series prediction.

The most critical task of deep learning applications is setting hyper-parameters and optimizing them. In order to effectively model a deep neural network, a large number of hyper-parameters need to be set. In experiments, the default parameters in Keras are used for deep neural network initialization (e.g., weight initialization). In order to avoid the over-fitting problem of the deep learning models, we apply several methods to solve it, such as a dropout policy with probability 0.3, which is used widely between layers (including convolutional layers, recurrent layers, and dense layers). And the default training parameters are: the batch size is 32, the epochs size is 100, and the lookup size is 1. We also use *tanh* as the activation function of the RNN model (include GRU and LSTM) and *ReLU* as the activation function of the CNN layers. In addition, we use Adam as an optimizer. The baseline model's network structure uses one hidden layer default and the number of neurons of each hidden layer is set to 128.

We use three convolution layers for local trend feature learning. Each layer has different filter size and kernel size parameter settings, say (64, 5), (32, 3), (16, 1). We use ReLU as the activation function. The bidirectional-LSTM layer is equipped using 128 hidden neurons for temporal features learning. We use mean square error (MSE) as the loss function of our DAQFF model. The activation function of the output layer is a linear function, which is also used for final prediction. Moreover, we apply min-max function to normalize the air quality time series data to [0,1]. Missing features in the experimental data are filled using the average value of the column in which they are located.

Additionally, for the Beijing PM2.5 Dataset experiment, we select the first four-year data for training and validation (three-year data for training, and the rest one-year data for validation) and select the last year data for testing (01/01/2014-12/31/2014). For the Urban Air Quality Dataset experiment, we select the first eight-month data for training (05/01/2014-12/31/2014) and select the last four-month data for testing (01/01/2015-04/30/2015). We use RMSE and MAE as the model error evaluation indicators, which are used to analyze the experimental results.

$$RMSE = \sqrt{\frac{1}{n}\sum_{i=1}^{n}(y_i - \widehat{y_i})^2} \tag{30}$$

$$MAE = \frac{1}{n}\sum_{i=1}^{n}|y_i - \widehat{y_i}| \tag{31}$$

where $\widehat{y_i}$ represents the predicted PM2.5 value, $y_i$ represents ground truth value and $n$ is the number of test dataset.

### 4.3 Single Step Forecasting Results Analysis

The single step PM2.5 prediction quantitative results of two real-world datasets are reported in Table 2, which give RMSE and MAE comparative analysis of ARIMA, SVR (rbf, linear and poly kernel), RNN, CNN, LSTM, GRU, and our proposed model DAQFF. As shown in Table 2, our model is superior to other baseline methods in terms of PM2.5 single-step forward prediction performance in both two datasets. Compared to the baseline shallow and deep learning models, our model reduces RMSE to 8.20 and MAE to 6.19 in Beijing PM2.5 Dataset, also has the lowest error in Urban Air Quality Dataset, which improves the forecasting accuracy obviously. In addition, the model error of classic deep learning models (such as LSTM, CNN, and GRU) are similar and also lower than shallow models. This means that deep learning models are more effective for air quality time series forecasting than traditional shallow learning models in single step prediction task.

TABLE 2
THE MODEL ERROR OF DAQFF AND COMPARISONS WITH OTHER BASELINE MODELS FOR THE SINGLE-STEP PM2.5 PREDICTION TASK.

| Models | Beijing PM2.5 Dataset | | Urban Air Quality Dataset | |
|---|---|---|---|---|
| | RMSE | MAE | RMSE | MAE |
| SVR-POLY | 42.61 | 31.82 | 56.35 | 47.20 |
| SVR-RBF | 41.86 | 34.93 | 50.51 | 42.26 |
| SVR-LINEAR | 30.60 | 20.47 | 29.23 | 18.82 |
| ARIMA | 24.52 | 12.50 | 27.92 | 14.35 |
| LSTM | 13.03 | 9.29 | 24.96 | 22.31 |
| GRU | 11.75 | 8.71 | 23.70 | 21.43 |
| CNN | 12.21 | 9.09 | 20.95 | 16.36 |
| RNN | 10.61 | 8.83 | 13.79 | 11.62 |
| **DAQFF** | **8.20** | **6.19** | **11.81** | **9.96** |

*Note: forward-step prediction size is 1, and model testing error (RMSE and MAE) are the prediction error of the next 1 hour (h1).*

Moreover, the prediction performance of baseline deep learning methods is dramatically better than the classic shallow learning methods such as SVR and ARIMA (1 to 2 times the gap). Our model performs the best since DAQFF can learn local trend features by one-dimensional CNN and long-term dependencies feature by Bi-directional LSTM of air quality multivariate time series data.

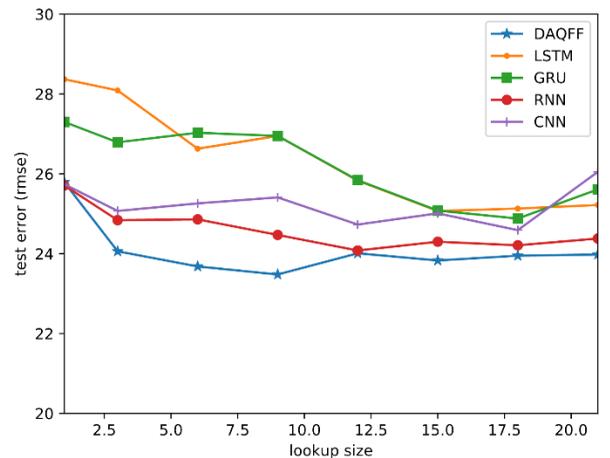

Fig. 5. RMSE of the proposed DAQFF model versus different lookup size and comparisons with baseline deep learning models in the experiment on Beijing PM2.5 Dataset. Hyper-parameters settings are: prediction size is 1 step, batch size is 64, the epoch is 30.

In addition, it is found by experiments that the choice of lookup size (lookup size is also called window size, which represents historical observations input size of the model) has an influence on the single step forecasting performance. We analyze the impact of lookup size among the baseline deep learning models and our model DAQFF in the Beijing PM2.5 Dataset. As Fig. 5 shows, we observe that compared to baseline deep models, our model DAQFF has the lowest prediction error versus different lookup sizes. With the increase of lookup size, these models error first decrease gradually. When the lookup size is around 9, the RMSE of DAQFF reaches the minimum. As the lookup size continues to increase, the model error remains stable or gradually increases, which may be led to overfitting problem.

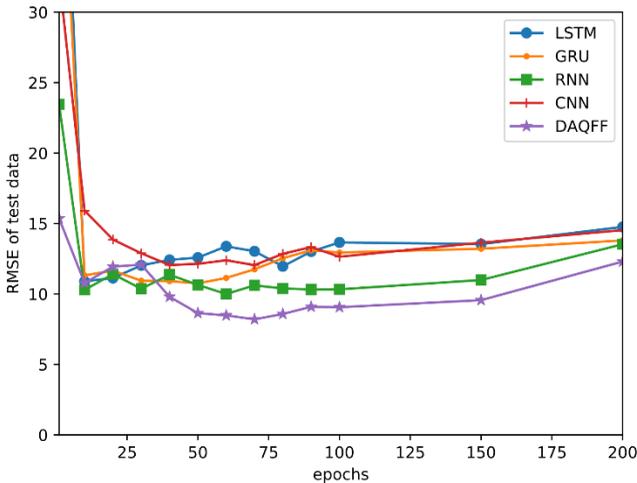

Fig. 6. RMSE of the proposed DAQFF model versus different epochs and comparisons with another baseline deep learning models in the experiment on Beijing PM2.5 Dataset. Hyper-parameters settings are: lookup and prediction size is 1, batch size is 64.

Then, we investigate the impact of epochs on different deep learning models. Fig. 6 shows the model error (RMSE) curve of the proposed DAQFF model versus different epochs and comparisons with another baseline deep learning models in the experiment on Beijing PM2.5 Dataset. It is obviously that our model DAQFF always maintains higher performance than the other baseline deep models versus different epochs. In addition, as the number of epochs increases, the prediction error of the deep learning models first gradually decreases. The RMSE of DAQFF achieves the lowest value when the epoch size is about 70 and then gradually grows when the epoch size continues to increase. It is clearly that the generalization capability does not improve obviously when the epoch size is larger than 70. Moreover, all models seem to be a little overfitting when the epoch size exceeds 90. In other words, the higher the epochs, the more the computational resources will be consumed. Although an increase of iterations can improve the training performance of the model, it will also cause overfitting problems.

In order to further evaluate the single step forecasting performance of DAQFF and baseline models in two real-world datasets, we analyze the PM2.5 prediction ability of DAQFF and the other two baseline models over the course of one month (31 days, 24 hours a day, all together including 744 observed PM2.5 data points). Figs. 7 (a) (b) (c) give a comparison of the ground truth (expected) and predicted one-step forward PM2.5 values of SVR, LSTM and DAQFF models in the experiment on Beijing PM2.5 Dataset. As shown in the figures, the performance of our model is better than those of SVR and LSTM models with single step forward prediction, especially in the time period of wave peak and trough of air quality time series data. Figs. 8 (a) (b) (c) show a comparison of the ground truth (expected) and predicted one-step forward PM2.5 values of SVR, LSTM and DAQFF models in the experiment on Urban Air Quality Dataset. Similarly, as shown in the figures, the single step forward prediction performance of our model is also better than those of SVR and LSTM models, both during the time period of wave peak and trough conditions.

In addition, we also observe that the single-step prediction performance of baseline models is sensitive to different data sets. For example, the SVR-RBF model has better forecasting performance in Beijing PM2.5 Dataset (see Fig. 7 (a)) than in Urban Air Quality Dataset (see Fig. 8 (a)). And the single-step forecasting performance of LSTM model is similar with SVR-RBF model, and the prediction performance of LSTM in Urban Air Quality Dataset (see Fig. 8 (b)) is worse than that in Beijing PM2.5 Dataset (see Fig. 7 (b)). But our model can maintain the best singe step prediction performance in both two datasets.

In summary, for single-step prediction of air quality time series under different experiment conditions, our model can maintain the best performance, and the prediction performance of the baseline deep learning models is also not bad, because the single-step prediction of time series is relative simple, which often only needs to follow the trend of the previous step to achieve a good forecasting performance. However, multi-step time series forecasting is not that simple, and it is often difficult to foresee what happens after multiple time steps. In the next section, we will analyze the performance of the multi-step forecasting models of air quality time series data.

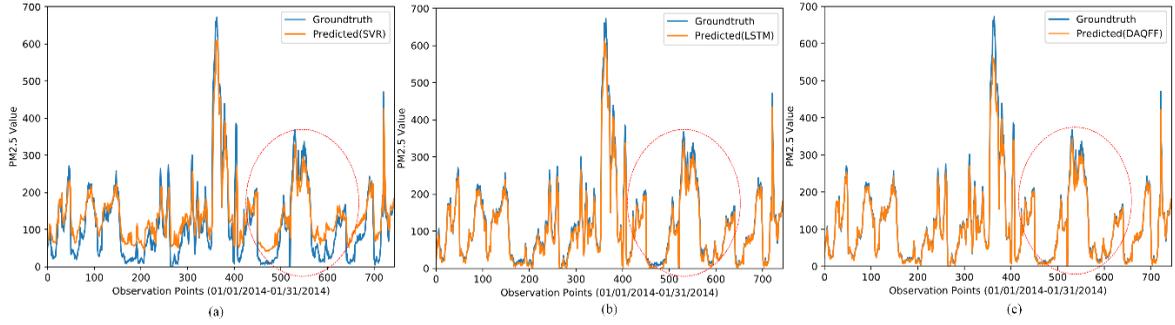

Fig. 7. In the experiment on Beijing PM2.5 Dataset, a comparison of single step ground truth and predicted PM2.5 value during one month (01/01/2014-01/31/2014) of different models (SVR, LSTM, and DAQFF). (a) SVR with RBF kernel model; (b) LSTM model; (c) DAQFF model.

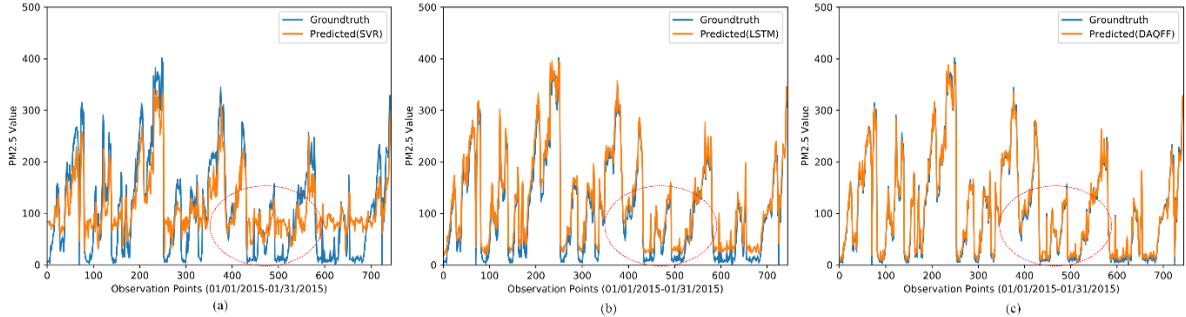

Fig. 8. In the experiment on Urban Air Quality Dataset, a comparison of single step ground truth and predicted PM2.5 value during one month (01/01/2015-01/31/2015) of 1001 station versus different models (SVR, LSTM, and DAQFF). (a) SVR with RBF kernel model; (b) LSTM model; (c) DAQFF model.

### 4.4 Multi-step Forecasting Results Analysis

The multi-step PM2.5 prediction quantitative results of two real-world datasets are reported in Table 3 (model testing error in the table is the average of the prediction error value in the next forward 6 hours, h1~h6), which gives RMSE and MAE comparative analysis of SVR (rbf, linear and poly kernel), RNN, CNN, LSTM, GRU, and our proposed model DAQFF. As shown in Table 3, our model is also superior to other methods in terms of PM2.5 multi-step prediction performance. Compared to the baseline shallow and deep learning models, our model reduces MAE to 27.53 in the Beijing PM2.5 Dataset, and also has the lowest MAE as 25.01 in the Urban Air Quality Dataset, which improves the forecasting accuracy obviously. It is worth noting that the testing error of classic deep learning models (RNN, CNN, LSTM, and GRU) are similar and larger than SVR-LINEAR model in the Beijing PM2.5 Dataset. Does this mean that the multi-step forecasting performance of the baseline deep learning model is worse than those of some shallow models (such as SVR-LINEAR)? In fact, it is not entirely true, as shown in Table 4, if long-term time step prediction is performed, we will find that the prediction performance of the baseline deep learning models will exceed the SVR-LINEAR model as the prediction time step increases. Taking the LSTM model as an example, in the next 3 hours (h1~h3), the average prediction error of the LSTM model is larger than that of the SVR-LINEAR model. However, when the forward prediction size is greater than 3, e.g. in the next h4~h6, h7~h12, or h13~h24 time period, the average prediction error of the LSTM model is lower than that of the SVR-LINEAR model. DAQFF model does not have this problem, since the performance of our model is better than that of the baseline models whether it is a short time step or a long time step prediction.

TABLE 3
THE MODEL ERROR OF DAQFF AND COMPARISONS WITH OTHER BASELINE MODELS FOR THE MULTI-STEP PM2.5 PREDICTION TASK.

| Models | Beijing PM2.5 Dataset | | Urban Air Quality Dataset | |
|---|---|---|---|---|
| | RMSE | MAE | RMSE | MAE |
| SVR-POLY | 56.62 | 44.94 | 64.02 | 50.82 |
| SVR-RBF | 57.66 | 46.32 | 65.11 | 53.59 |
| SVR-LINEAR | 49.82 | 36.82 | 53.48 | 36.35 |
| LSTM | 57.49 | 44.12 | 58.25 | 44.28 |
| GRU | 52.61 | 38.99 | 60.76 | 45.53 |
| RNN | 57.38 | 44.69 | 60.71 | 46.16 |
| CNN | 52.85 | 39.68 | 53.38 | 38.21 |
| **DAQFF** | **43.49** | **27.53** | **46.49** | **25.01** |

*Note: forward multi-step prediction size is **6**, and model testing error (RMSE and MAE) are the average of the prediction error in the next forward 6 hours (h1~h6).*

Next, we analyze the impact of forward prediction size among the baseline deep learning models and DAQFF. As shown in Table 4, in the Beijing PM2.5 Dataset, the performance of PM2.5 multi-step forward prediction is significantly lower than that of single step forward prediction (see Table 2). As the forward prediction size increases, the forecasting performances of these models gradually decrease. But we can observe that compared to baseline methods, our DAQFF model also has the lowest prediction error (RMSE and MAE) versus different forward prediction sizes.

TABLE 4
IN THE EXPERIMENT ON BEIJING PM2.5 DATASET, THE MODEL ERROR OF DAQFF AND COMPARISONS WITH OTHER BASELINE MODELS FOR THE MULTI-STEP PREDICTION OF PM2.5 VALUES IN THE NEXT 24 HOURS.

| Models | RMSE | | | | MAE | | | |
|---|---|---|---|---|---|---|---|---|
| | 1h~3h | 4h~6h | 7h~12h | 13h~24h | 1h~3h | 4h~6h | 7h~12h | 13h~24h |
| SVR-POLY | 48.99 | 64.26 | 75.70 | 84.91 | 39.14 | 50.74 | 59.71 | 66.92 |
| SVR-RBF | 51.15 | 64.18 | 75.46 | 84.92 | 41.76 | 50.88 | 59.57 | 67.02 |
| SVR-LINEAR | 38.69 | 60.96 | 76.24 | 85.60 | 27.31 | 46.32 | 60.10 | 67.04 |
| RNN | 49.50 | 65.27 | 77.06 | 80.56 | 38.89 | 50.49 | 59.88 | 59.18 |
| CNN | 45.95 | 59.76 | 70.83 | 79.18 | 35.62 | 43.74 | 51.21 | 57.91 |
| LSTM | 45.88 | 57.51 | 69.52 | 79.15 | 35.31 | 40.32 | 48.76 | 57.17 |
| GRU | 47.32 | 57.90 | 69.57 | 79.23 | 37.01 | 40.98 | 48.77 | 57.25 |
| **DAQFF** | **34.35** | **52.64** | **66.14** | **77.38** | **20.91** | **34.15** | **44.93** | **54.58** |

*Note: Hyper-parameters settings are: forward multi-step prediction size is 24, lookup size is 1, epochs is 100, batch-size is 64, and testing error (RMSE, MAE) of t~t+n is the average of the prediction error in the next forward n hours.*

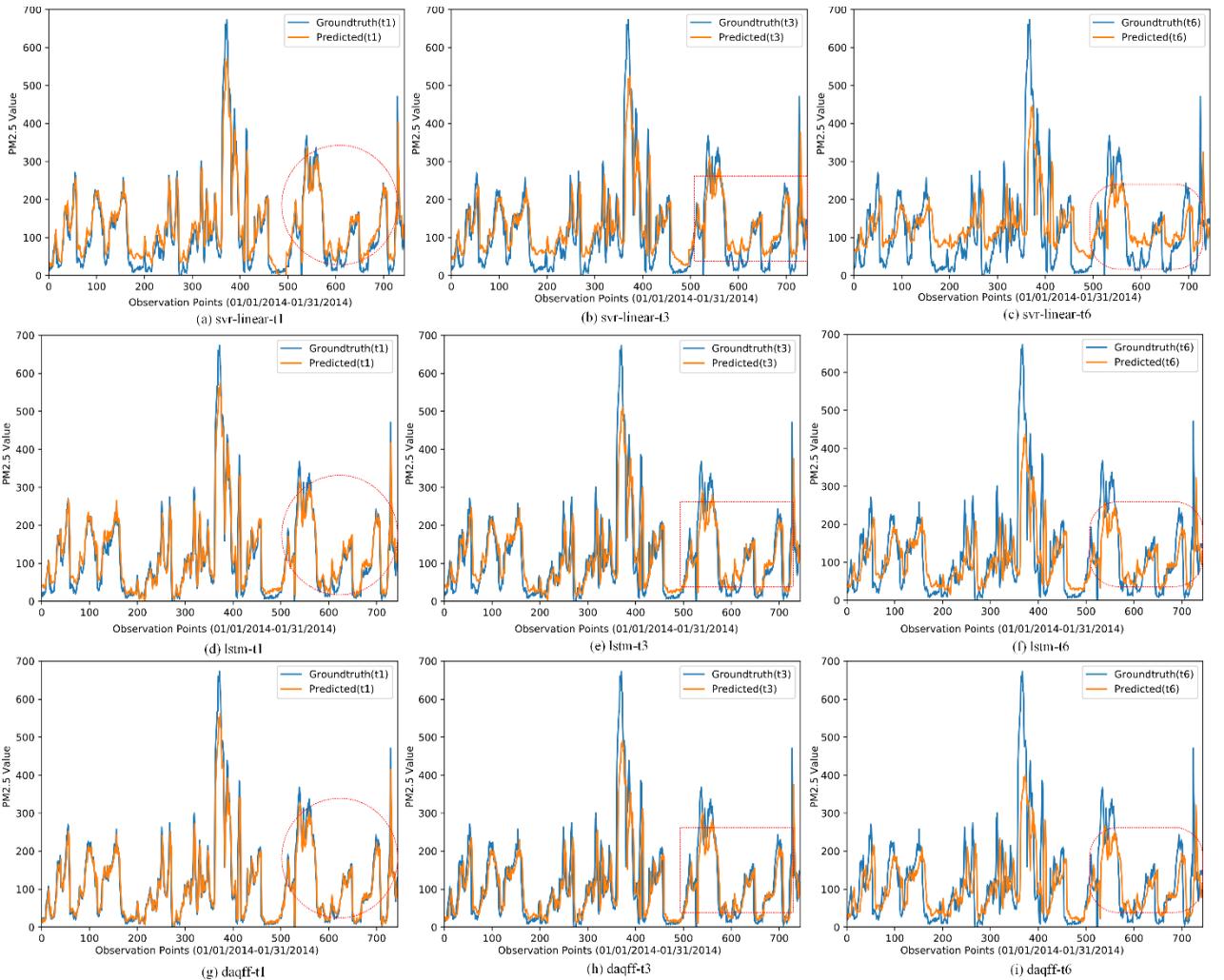

Fig. 9. In the experiment on Beijing PM2.5 Dataset, a comparison of multi-step (next *t*1, *t*3, and *t*6) ground truth and predicted PM2.5 value during one month (01/01/2014-01/31/2014) of different models (SVR, LSTM and DAQFF). (a) SVR-LINEAR model for next 1 hour (*t*1) prediction; (b) SVR-LINEAR model for the third hour of the future (*t*3) prediction; (c) SVR-LINEAR model for the 6th hour in the future (*t*6) prediction; (d) LSTM model for next 1 hour (*t*1) prediction; (e) LSTM model for the third hour of the future (*t*3) prediction; (f) LSTM model for the 6th hour in the future (*t*6) prediction; (g) DAQFF model for next 1 hour (*t*1) prediction; (h) DAQFF model for the third hour of the future (*t*3) prediction; (i) DAQFF model for the 6th hour in the future (*t*6) prediction;

TABLE 5
IN THE EXPERIMENT ON URBAN AIR QUALITY DATASET, THE MODEL ERROR OF DAQFF AND COMPARISONS WITH OTHER BASELINE MODELS FOR THE MULTI-STEP PREDICTION OF PM2.5 VALUES IN THE NEXT 48 HOURS.

| Models | RMSE | | | | MAE | | | |
|---|---|---|---|---|---|---|---|---|
| | 1h~6h | 7h~12h | 13h~24h | 25h~48h | 1h~6h | 7h~12h | 13h~24h | 25h~48h |
| SVR-POLY | 64.02 | 83.72 | 105.73 | 109.98 | 50.82 | 64.01 | 83.96 | 86.83 |
| SVR-RBF | 65.11 | 83.96 | 88.81 | 90.38 | 53.59 | 67.52 | 70.32 | 74.23 |
| SVR-LINEAR | 53.48 | 83.85 | 91.06 | 93.11 | 36.35 | 68.01 | 73.72 | 72.73 |
| RNN | 60.71 | 86.25 | 100.54 | 115.13 | 46.16 | 67.94 | 81.97 | 99.45 |
| GRU | 60.76 | 95.59 | 107.82 | 111.06 | 45.53 | 72.81 | 82.45 | 90.25 |
| LSTM | 58.25 | 88.52 | 96.61 | 103.21 | 44.28 | 69.39 | 76.40 | 84.65 |
| CNN | 53.38 | 83.48 | 94.85 | 94.53 | 38.21 | 61.97 | 71.94 | 77.92 |
| **DAQFF** | **46.49** | **69.15** | **77.88** | **80.06** | **25.01** | **48.37** | **59.69** | **61.75** |

*Note: Hyper-parameters settings are: forward multi-step prediction size is 48, lookup size is 1, epochs is 100, batch-size is 64, and testing error (RMSE, MAE) of t~t+n is the average of the prediction error in the next forward n hours.*

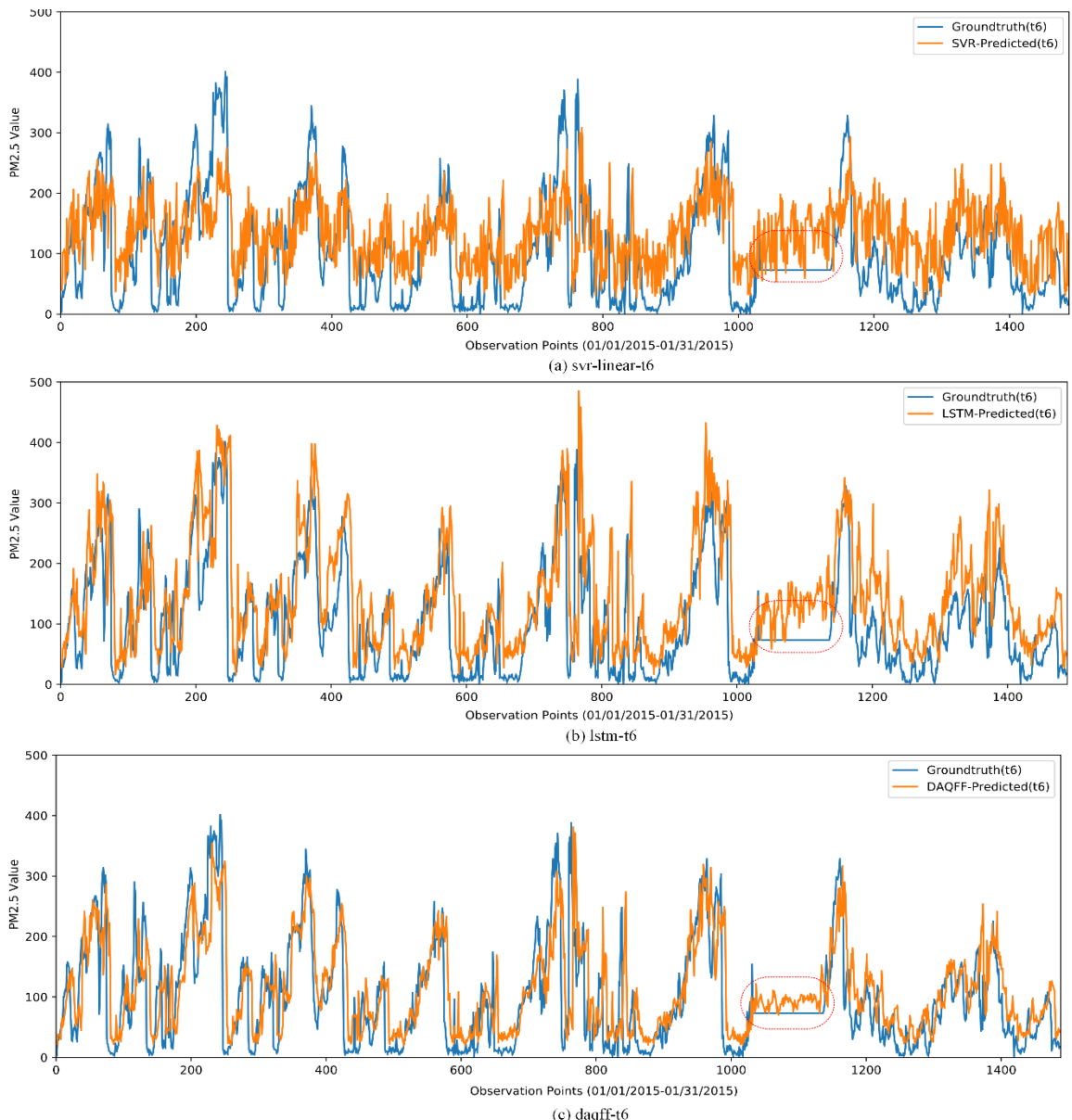

Fig. 10. In the experiment on Urban Air Quality Dataset, forecast the PM2.5 value of one station (no.1001) based on the trained model of 36 stations data. A comparison of multi-step (*t*6) ground truth and predicted PM2.5 value during two months (01/01/2015-02/28/2015) of SVR-LINEAR, LSTM and DAQFF model (a) SVR-LINEAR for the next 6th hour (*t*6) prediction; (b) LSTM for the next 6th hour (*t*6) prediction; (c) DAQFF for the next 6th hour (*t*6) prediction.

In order to further analyze and compare the forecasting performance of DAQFF and the other baseline models, we analyze the multi-step forecasting ability of our model in the experiment on Beijing PM2.5 Dataset under different time-step (*t*1, *t*3 and *t*6) forward prediction over the course of test data in one month (24 hours per day, 31 days, totally 744 time-step points, 01/01/2014-01/31/2014). Figs. 9 (a)-(i) give a comparison of the ground truth and the predicted PM2.5 value of the experiment with different models (SVR-LINEAR, LSTM, and DAQFF) under different predict sizes' (1 time-step, 3 time-step, and 6 time-step) conditions, where x-coordinate indicates the observation time-steps and y-coordinate indicates the PM2.5 value. As shown in these figures, the multi-step forecasting performance of our model is better than those of SVR-LINEAR represented shallow models and LSTM represented deep learning models, not only under short prediction size but also under long prediction size conditions, especially in the time period of wave peak and trough of test data. Moreover, as the prediction time-steps grow, the predictive performance of SVR models decreases dramatically, but our DAQFF model can maintain the best performance.

Moreover, we analyze whether DAQFF model can maintain the same forecasting ability for different air quality data sets. In the experiment on Urban Air Quality Dataset, we further verify the forecasting performance of DAQFF model. As Table 5 shows, compared with other models, the forecasting performances of SVR-POLY, RNN, and GRU model have large fluctuations in long-term time step (e.g. h13~h24, h25~h48) prediction. In addition, what is interesting about the data in Table 5 is that as the prediction time step grows, our model can maintain a significant improvement over the baseline models compared with the data of Table 4. In short, our model can maintain optimal prediction performance over short-term or long-term time step forecasting conditions.

Figs. 10 (a) (b) (c) give a zoom in a comparison of the ground truth (expected) and six-step forward predicted PM2.5 values of SVR-LINEAR, LSTM, and DAQFF models. Through the comparison of these three figures, it is found that the multi-step forecasting performance of classic deep learning model like LSTM is better than that of shallow model SVR-LINEAR, and SVR-LINEAR model cannot effectively predict the PM2.5 values such as the wave valley and the wave peak of air quality time series data when the prediction time step is 6. Although the prediction performance of LSTM is not bad, it is not accurate enough to predict the wave valley and wave peak values of PM2.5 time series. DAQFF has better performance at all time period observation points and can effectively predict PM2.5 values under different conditions (e.g. in the case of missing values, as shown in the red wireframe section in Fig. 10). In short, through experimental results, we find that the overall multi-step prediction performance of DAQFF is the best, no matter if it is on general weather day or extreme weather day, weekdays or weekends.

Finally, we have compared the multi-step forecasting performance of DAQFF and the state-of-the-art method, Zheng et al. [19], which used the same Urban Air Quality Dataset. As shown in Table 6, compared with the method of Zheng et al. [19], when the prediction size is less than 6, the forecasting performance of the DAQFF model is worse than that of the method of Zheng et al., but for long time-step forward forecasting, our model DAQFF performs better. It should be noted that the benchmarks are not consistent for comparison of different studies due to the small number of benchmark data sets available in the air quality forecasting field, even the released Urban Air Quality Dataset [19] having not been well pre-processed (e.g. there are many missing values to which different processing methods can be applied).

TABLE 6
COMPARISONS AMONG DIFFERENT RESEARCHES WITH URBAN AIR QUALITY DATASET

| Models | MAE | | | |
|---|---|---|---|---|
| | 1h~6h | 7h~12h | 13h~24h | 25h~48h |
| Zheng et.al [19] | **23.70** | 52.40 | 63.90 | 69.0 |
| **DAQFF** | 25.01 | **48.37** | **59.69** | **61.75** |

All in all, for the proposed DAQFF, the PM2.5 prediction can be well matched with the ground truth with single step forward prediction, also has better performance than baseline models with multi-step forward prediction, which implies the deep air quality forecasting framework can effectively learn the local trend and long-term temporal dependence characteristics of multivariate air quality time series data. The proposed air quality forecasting model DAQFF which is based on a hybrid deep learning structure can provide a useful reference for air pollution management and early warning.

## 5 CONCLUSION AND FUTURE WORK

In this paper, we proposed a new air quality forecasting framework (DAQFF) for PM2.5 single step forward and multi-step forward prediction, which is based on a hybrid deep learning method. DAQFF consists of two deep neural networks: one-dimensional CNNs and Bi-directional LSTM. It can learn the correlation features of local trend and spatial-temporal dependencies pattern of multivariate air quality related time series data. Experiments showed that the proposed model has better performance than classic shallow learning and deep learning models, which can explore and learn the interdependence and nonlinear correlations of multivariate air quality related time series (e.g. temperature, humidity, wind speed, SO2, PM10 and PM2.5 itself) effectively. The main contributions of this paper are as follows:

1) We firstly proposed a new hybrid deep learning framework which can deal with hierarchical feature representation and multi-scale spatial-temporal dependency fusion learning in an end-to-end process for air quality forecasting.

2) This study was the first attempt to combine multiple one-dimensional CNNs and bi-directional LSTM for hybrid fusion learning of air quality related multivariate time series data, which can extract spatial-temporal dependency and correlation features for air quality multi-step forecasting modeling.

3) We demonstrated the effectiveness of our model by testing it on two real-world air quality datasets, and the experimental results indicated that our model has good forecasting ability (not only single step but also multi-step forecasting). It was also showed that the proposed model has better prediction ability than typical shallow learning and baseline deep learning models.

In future research, we believe that the abrupt change (also called outlier or anomaly point) of air pollution time series deserves further study. If we can predict the sudden change of air pollution in advance, which will greatly improve the multi-step forecasting ability of our model. In addition, the model DAQFF also needs to be researched in depth and improved under different forecasting conditions.

## ACKNOWLEDGMENT

This research was partially supported by the National Natural Science Foundation of China (Nos. 61773324, 61572407), the "Center for Cyber-physical System Innovation" from The Featured Areas Research Center Program within the framework of the Higher Education Sprout Project by the Ministry of Education (MOE) in Taiwan and MOST under 106-2221-E-011-149-MY2 and108-2218-E-011-006.